\documentclass{article}

\usepackage{hyperref}
\usepackage{url}
\usepackage[pdftex]{graphicx}
\usepackage{booktabs}
\usepackage{multirow}
\usepackage{tikz}
\usepackage[flushleft]{threeparttable}

\usepackage[accepted]{icml2024}

\usepackage{amsmath}
\usepackage{amssymb}
\usepackage{mathtools}
\usepackage{amsthm}

\usepackage[capitalize,noabbrev]{cleveref}

\theoremstyle{plain}

\theoremstyle{definition}

\theoremstyle{remark}

\def\name{$R^2$ loss}
\def\fname{Range Restriction Loss}

\icmltitlerunning{R2 Loss: Range Restriction Loss for Model Compression and Quantization}

\begin{document}

\twocolumn[
\icmltitle{$R^2$ Loss: Range Restriction Loss for Model Compression and Quantization}



\icmlsetsymbol{equal}{*}

\begin{icmlauthorlist}
\icmlauthor{Arnav Kundu}{equal}
\icmlauthor{Chungkuk Yoo}{equal}
\icmlauthor{Srijan Mishra}{}
\icmlauthor{Minsik Cho}{}
\icmlauthor{Saurabh Adya}{}
\end{icmlauthorlist}

\centering{Apple Inc.}

\icmlcorrespondingauthor{Arnav Kundu}{a_kundu@apple.com}
\icmlcorrespondingauthor{Chungkuk Yoo}{ckyoo@apple.com}

\icmlkeywords{Machine Learning, ICML}

\vskip 0.3in
]


\begin{abstract}
Model quantization and compression is widely used techniques to reduce usage of computing resource at inference time. While state-of-the-art works have been achieved reasonable accuracy with higher bit such as 4bit or 8bit, but still it is challenging to quantize/compress a model further, e.g., 1bit or 2bit. To overcome the challenge, we focus on outliers in weights of a pre-trained model which disrupt effective lower bit quantization and compression. 
In this work, we propose \fname~(\name) for building lower bit quantization and compression friendly models by removing outliers from weights during pre-training. By effectively restricting range of weights, we mold the overall distribution into a tight shape to ensure high quantization bit resolution, therefore allowing model compression and quantization techniques can to utilize their limited numeric representation powers better. We introduce three different, $L_\infty$ \name, its extension Margin \name~ and a new Soft-Min-Max \name~to be used as an auxiliary loss during full-precision model training. These \name~can be used in different cases such as $L_\infty$ and Margin \name~would be effective for symmetric quantization, while Soft-Min-Max \name~shows better performance for model compression. In our experiment, \name~improves lower bit quantization accuracy with state-of-the-art post-training quantization (PTQ), quantization-aware training (QAT), and model compression techniques. With \name, MobileNet-V2 2bit weight and 8bit activation PTQ, MobileNet-V1 2bit weight and activation QAT, ResNet18 1bit weight compression are improved to 59.49\% from 50.66\%, 59.05\% from 55.96\%, and 52.58\% from 45.54\%, respectively.


\end{abstract}

\section{Introduction}
\begin{figure*}[!h]
\centering
\begin{minipage}{0.5\textwidth}
  \centering
  \includegraphics[width=1\linewidth]{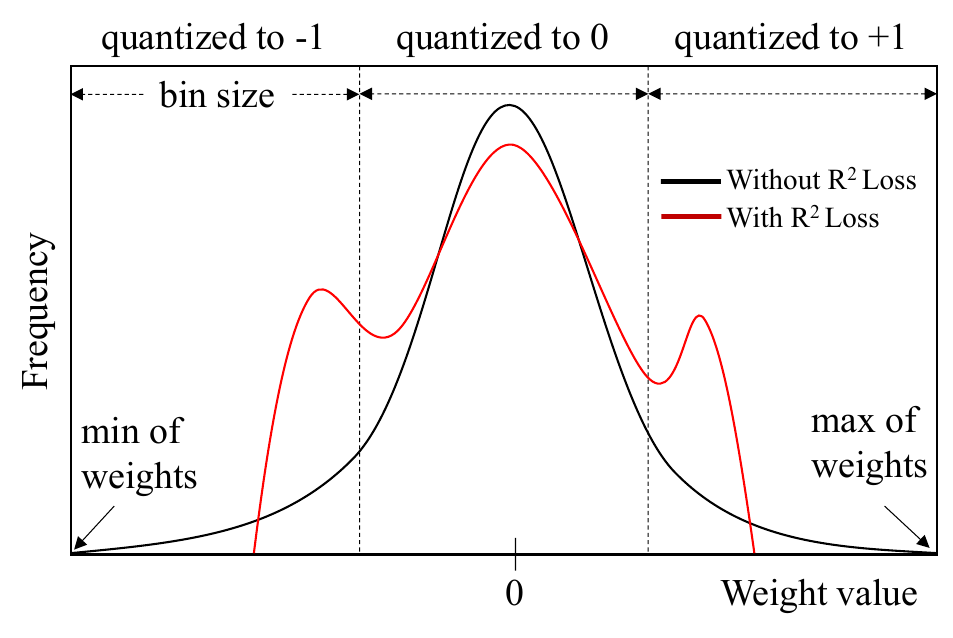}
  
\end{minipage}%
\begin{minipage}{.5\textwidth}
  \centering
  \includegraphics[width=1\linewidth]{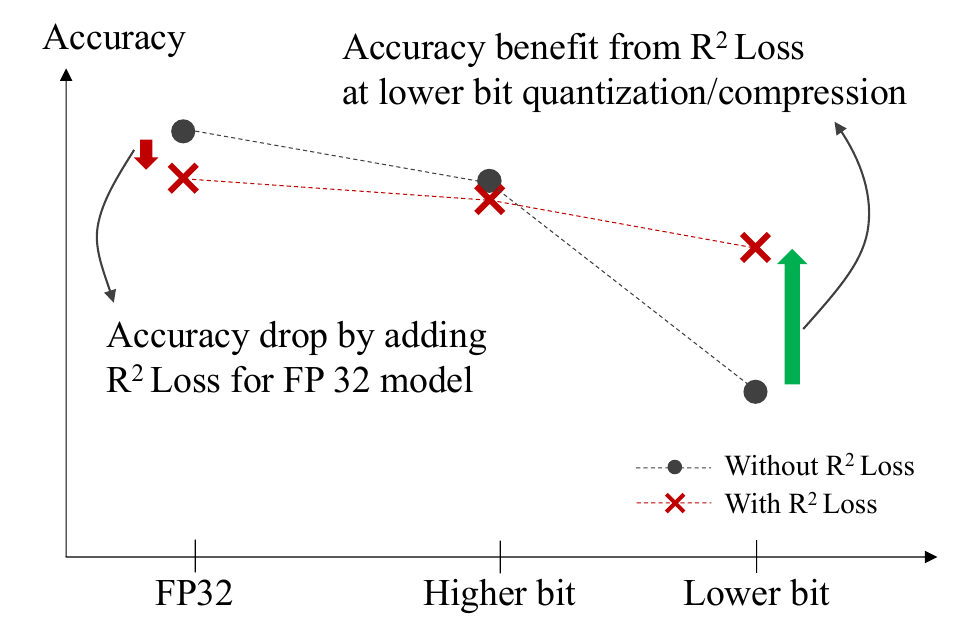}
\end{minipage}
\vskip -0.2in
\caption{(Left) Example weight distribution for weight quantization with three bins. (Right) Expected benefit of \name~for low bit quantization/compression.}
\label{fig_weight_dist}
\vskip -0.1in
\end{figure*}

Deep neural networks have become popular in human computer interaction, anomaly detection, financial markets, etc. Since a lot of applications running these models run on edge devices, running these compute-intensive models requires a balance of accuracy, latency, power efficiency and size for these models. 

Quantization is an effective way of reducing the power, latency, and size of neural networks. This requires that these quantized models are executed on specialized platforms with low-bit supports and at the cost of accuracy drop \cite{hawq,haq}. Post-training quantization involves distributing all the available weights in a layer into bins spread equally apart across the range. Quantization-aware training techniques ~\cite{bengio2013estimating,zhou2016dorefa}, use stochastic gradient descent to quantize and optimize the weights (i.e., mapping each floating point number to a set of discrete bins according to the precision target). Quantization bit-resolution is inversely proportional to the range of weights and affects accuracy of the quantized models. Since outliers tend to increase range, outliers are detrimental for quantization friendly models. 

As an  example, lets assume we want to quantize the weight distributions shown in \cref{fig_weight_dist} (left) into 3 bins. For the original distribution in black most of the weights will be quantized to zero and the model accuracy would drop significantly. This problem gets worse for low bit quantization and compression such as one or two bit quantization.






We introduce \textbf{\fname~(\name)}, a simple yet powerful method that helps to remove outliers during training without severely affecting full precision accuracy and provides a quantization or compression friendly checkpoint. Using \name~we intend to trim the edges of the black distribution and convert it to a distribution similar to the one shown in red in \cref{fig_weight_dist} (left). The expected behavior of \name~is that it possibly would regress full-precision model's accuracy slightly as it works as additional weight regularization like weight decay. However, the full-precision model trained with \name~would have a quantization/compression friendly weight distribution removing outlier weights, so lower bit quantization accuracy can be improved as shown in \cref{fig_weight_dist} (right). In case of higher bit quantization such as 4bit or 8bit, a model might already have enough bits to properly represent weights even including outliers. Therefore the benefit of \name~could be limited. Also as state-of-the-art techniques show reasonable accuracy at higher bit quantization at marginal accuracy regression from full-precision models, there is not much room for improvement at higher bit quantization/compression.

We propose three different formulations of \name~and through experiments show that models trained with them are more quantization and compression friendly. $L_\infty$~\name~is simple and intuitive to penalize the outlier weights during pre-training. Margin \name~is an extension of $L_\infty$. It penalizes weights which are larger than a margin, while minimizing the width of the margin. Soft-Min-Max \name~smoothly penalizes not only outliers but also near-
outlier weights. These three \name~can be used in different cases considering their characteristics. $L_\infty$ and Margin \name~would be more effective to symmetric quantization as they push to a symmetric weight distribution. Soft-Min-Max \name~could be better for compression than others as it makes asymmetric weight distribution and smoothly regularizes weights with a differentiable parameter.

We show that \name~works well with state-of-the-art quantization and compression algorithms by conducting accuracy evaluation with many different quantization and compression approaches such as DFQ\cite{nagel2019data}, AdaRound\cite{nagel2020up}, SQuant\cite{guo2022squant}, PD-Quant\cite{liu2023pd}, PACT\cite{choi2018pact}, LSQ\cite{esser2019learned}, EWGS\cite{lee2021network}, and DKM\cite{cho2021dkm} with various models like MobileNet-V1, MobileNet-V2, ResNet18, ResNet50, ResNet101, and MobileBERT\cite{sun2020mobilebert}.


Our experiments on MobileNet-V1 \cite{howard2017mobilenets} and MobileNet-V2 \cite{sandler2018mobilenetv2} using post training quantization techniques like DFQ \cite{nagel2019data} shows $>$ 10\% improvement using \name~trained models. Even with newer methods like SQuant \cite{guo2022squant} we observe an improvement of $>$ 5\%. \name~ also shows similar gain in accuracy for PTQ methods (Adaround, SQuant, PD-Quant \cite{nagel2020up,guo2022squant,liu2023pd}) for larger models like ResNet-50 and ResNet-101 \cite{he2016deep}. For QAT methods like EWGS \cite{lee2021network} $R^2$ trained models are roughly 4\% better than the ones trained without \name~for 2bit weight and activation quantization. For models compressed using 32x compression, \name~improves the accuracy of parameter constrained models like MobileNet-V1 by 5\% (2-bit,2-dim). We have also extended  \name~to fine-tuning MobileBERT on QNLI task, where we see an absolute 2\% improvement in  accuracies for 1-bit and 2-bit compression.

\section{Related Works}
\label{related_works}

\subsection{Model compression}

The simplest and one of the most effective form of compression involves sharing weights within a layer. Deep Compression \cite{HanMD15} introduced k-means clustering based weight sharing for compression. Initially, all weights belonging to the same cluster, share weight of the cluster centroid. During forward pass, loss is calculated using the shared weights which are then updated during backward pass. This leads to a loss of accuracy and model train-ability because the weight to cluster assignment is intractable during weight update \cite{yinl2019}. DKM \cite{cho2021dkm} introduces differentiable k-means clustering, therefore making cluster assignments tractable. During forward clustered weights are used, however during backward the gradient is applied on the original weights.

\subsection{Model quantization}
Model quantization reduces the memory footprint of a model by reducing the representative bits per weight for a given model. It also quantizes activation values so that we can convert floating point computation into integer computation which gives us a benefit of hardware efficiency. In this paper we have applied our \name~with various training time quantization (quantization-aware training, QAT) algorithms like EWGS \cite{lee2021network}, LSQ \cite{esser2019learned} and DoReFa \cite{zhou2016dorefa} used in PACT \cite{choi2018pact}. PACT clips activation values with a trainable parameter for activation quantization and uses DoReFa for weight quantization. LSQ quantizes weights and activations with learnable step size (scale or bin size). EWGS applies gradient scaling with respect to position difference in between original full precision weights and quantized weights based on LSQ.

Also, we have compared our \fname~with a state-of-the-art post-training quantization (PTQ) methods. DFQ \cite{nagel2019data} equalizes per-channel weight ranges by applying per-channel scaling factors. It resolves the wide weight range problem across channels, but still the weight range would remain wide for lower bit quantization like 4bit as DFQ does not target outliers within a channel. In our experiment, models trained with our \fname~can be effectively quantized to 4bit weight / 8bit activation by PTQ without DFQ. AdaRound \cite{nagel2020up} proposed adaptive rounding for quantization bin assignment instead of nearest rounding. Pre-trained models with \fname~also show better quantization accuracies with AdaRound than models trained with just L2 norm which is well-known regularization. SQuant \cite{guo2022squant} decomposes a layer by the Hessian-based optimization objective into three diagonal sub-items, element-wise, kernel-wise, and output channel-wise. It then compose the sub-items in a quantized domain with respect to constrained absolute some of error. PD-Quant \cite{liu2023pd} quantizes weights by comparing model prediction result before and after quantization of each layer.

In our extensible experiments, we show our \fname~improves accuracies with cutting-edge QAT and PTQ for lower bit quantization like 2bit weight / 2bit activation and 4bit weight / 8bit activation.

\subsection{Regularization for quantization}
Regularization is a well-known technique for over-fitting. But some research works used the regularization concept in the context of quantization. \cite{kure} show that uniform distribution of weights are more robust to quantization than normally-distributed weights. To this they propose KURE (KUrtosis REgularization) to minimize the kurtosis of weights and ensure a uniform distribution. This method is independent of the quantization bit-width, therefore supports PTQ (Post-Training Quantization) in addition to QAT (Quantization Aware Training). However, this method is best suited for generic models which need to be specifically tuned for a given bit precision use case. To reduce the accuracy drop due to quantization \cite{binregularization} proposes to constrain weights to predefined bins based on the quantization bit-width. However, selecting these bins is a difficult process and the output of the models in very sensitive to the bin selection. In addition to that these methods ignore the effect of quantizing the first and last layers of the models.

A key difference between \name~and the existing regularization research for quantization is that other works does not explicitly consider the outliers nor weight ranges in a pre-trained model. Therefore, as shown in \cref{fig:weight_dist_mnv2}, the model trained with KURE still have wider weight ranges than models trained with \name~so that a quantization model from KURE would have a larger bin size which leads to bigger quantization error. Therefore, in our comparison experiment, our \name~shows noticeable improvement from KURE. 
\section{\fname}

\begin{figure}[t!]
\vskip 0.0in
\begin{center}
\includegraphics[width=1\linewidth]{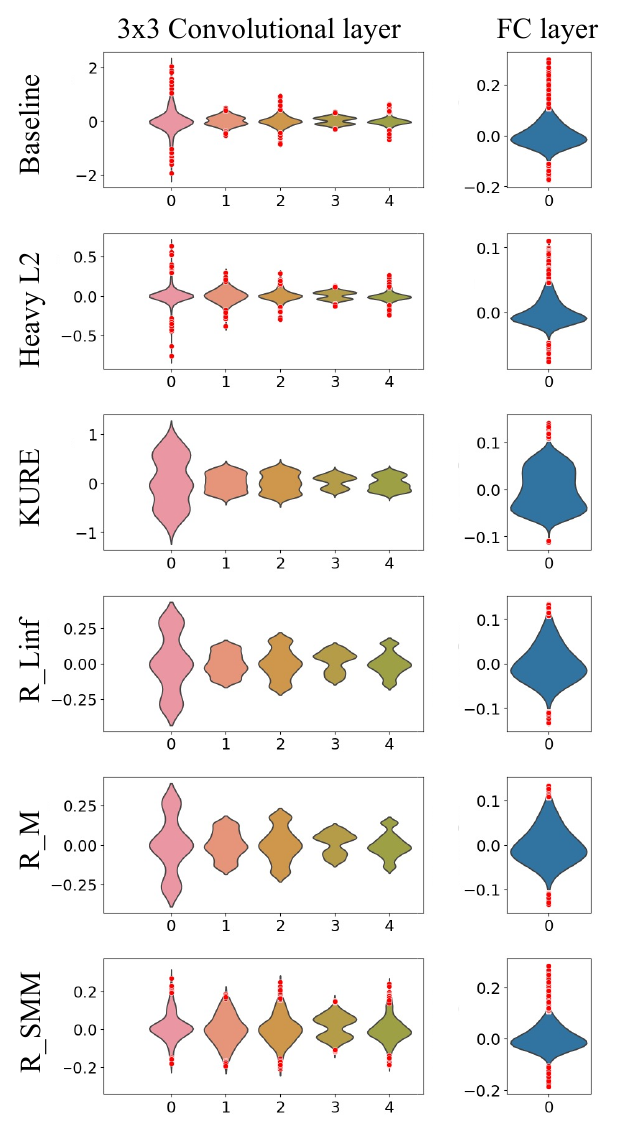}
\vskip -0.2in
\caption{Weight distribution of the first five 3x3 convolution layers and FC layer of MobileNet-V2 using L2 norm (baseline), heavy L2 norm (10x heavy L2 norm than baseline) and the proposed \fname~(the red dots correspond outliers). KURE: Kurtosis Regularization~\cite{kure}.}
\label{fig:weight_dist_mnv2}
\end{center}
\vskip -0.0in
\end{figure}

We introduce \fname~as an auxiliary loss to reduce the range of weights for every layer to get better pre-trained models for further quantization or compression. Just like $L_1$ and $L_2$ regularization our approach is invariant to the quantization or compression technique used. But as opposed to $L_1$ or $L_2$ regularization, \fname~only affects the outliers in weights by penalizing them while maintaining accuracy of the full precision model. We demonstrate that $L_2$ regularization (1x(baseline) and 10x(heavy L2)) does not solve the problem of outliers in \cref{fig:weight_dist_mnv2}. As a reference for the expected weight distribution for a quantization friendly model we use the weight distribution from a model trained using KURE \cite{kure}. While, the idea of minimizing range can be formulated in various ways we propose 3 different ways of defining $R^2$. We start from $L_\infty$ loss, extend it to margin loss and finally introduce soft-min-max loss for adding \name~to the training loss.

    \textbf{$L_\infty$ \name}: This method tries to penalize only the outliers in an iterative manner during training by adding $L_{\infty}(W)$ as an auxiliary loss for every layer in the model.
    \begin{equation}
         L_{reg}=\sum L_{\infty}(W)
    \end{equation}
    The effect of this formulation is described in \cref{fig:weight_dist_mnv2} where it shows that \fname~helps to get rid of all outliers in the model. In addition, it brings the overall range of weight down in contrast to KURE \cite{kure}, while, also making the weight distribution similar to a mixture of Gaussians as seen in KURE.
    
    \textbf{Margin \name}: This is an extension of $L_{\infty}(W)$ \name, where, we define a margin for the range of allowed weights. Any weight outside this margin is penalized. In addition, we also penalize the width of the margin to ensure that the range of the overall weight distribution is small. The auxiliary loss for a given weight W is shown in \cref{eq:margin}. Here $M$ is a learnable parameter per layer.
    \begin{equation}
        L_{reg}=\sum (|M| + \max(|W|-|M|,0))
        \label{eq:margin}
    \end{equation}
    The effect of margin \name~is similar to that of $L_\infty$ in terms of the final weight distribution as evident from \cref{fig:weight_dist_mnv2}. The only difference is that margin \name~penalizes all weight outside the margin per iteration in contrast to penalizing only the maximum weight.
    
    \textbf{Soft-min-max \name \label{sec:smm}}: In this approach, we propose an asymmetric \name, to eliminate the constraint on the magnitude of weights and strictly enforce it on the range of weights. We hypothesize that such a technique will improve asymmetrically quantized/compressed models using techniques such as DKM \cite{cho2021dkm}. The loss for a given weight W is described in \cref{eq:sminmax}. 
    \begin{equation}
    \begin{split}
        s_{max} &= \frac{\Sigma (W \odot e^{\alpha \times (W-W_{max})})}{\Sigma e^{\alpha \times (W-W_{max})}} \\
        s_{min} &= \frac{\Sigma (W \odot e^{-\alpha \times (W-W_{min})})}{\Sigma e^{-\alpha \times (W-W_{min})}} \\
        L_{reg} &= (s_{max}-s_{min}) + e^{-\alpha}
    \end{split}
    \label{eq:sminmax}
    \end{equation}
    Here temperature $\alpha$ is a learnable parameter per layer. $e^{-\alpha}$ term in the auxiliary loss $L_{reg}$, encourages temperature $\alpha$ to increase during training time optimization process to approach hard-min-max loss towards the end of training. This loss smoothly penalizes not only outliers but also near-outlier weights together rather strictly brings only outliers down like other \name es. Therefore, it might be susceptible to outliers as seen in \cref{fig:weight_dist_mnv2}.

All the above mentioned \name~were employed during training time of the base model itself and not during quantization or compression. This was done because the purpose of \name~is to provide effective initial weights for compression or quantization. This ensures extensibility of \name~to any quantization or compression technique.
\section{Experiment}
\label{experiment}

\subsection{Experiment settings}
\subsubsection{Pre-training from scratch with and without \name}
We train ResNet-18~\cite{he2016deep}, MobileNet-V1~\cite{howard2017mobilenets} and MobileNet-V2~\cite{sandler2018mobilenetv2} on ImageNet 1K~\cite{deng2009imagenet} with proposed \fname~on a x86 Linux machine with eight GPUs to get pre-trained models before model compression and quantization-aware training. We set initial learning rates to 1.0, 0.4 and 0.4 for ResNet-18, MobileNet-V1 and MobileNet-V2 respectively. We use SGD with 0.9 of momentum with Nesterov. We apply 1e-4 of weight decay (L2 norm weight regularization) for ResNet-18 and 4e-5 for MobileNet-V1 and V2. For heavy L2-regularization, in ~\cref{fig:weight_dist_mnv2}, we use 4e-4 of weight decay (10x heavier than baseline) for MobileNet-V2 to see whether heavy L2-regularization helps quantization or not as a naive solution for range restriction. Strength of \name~is set to 0.01. For Margin \name, the margin threshold is initialized with 2x the standard deviation of the initialized weights. In Soft-min-max \name~training, the learnable parameter $\alpha$ is initially set to 0.1. For comparison, we use pre-trained models of Resnet-18 from Torchvision. As we are using modified version of ResNet-50, and ResNet-101, MobileNet-V1 and V2 for better FP32 performance, we trained those models from scratch without \name~using the same settings above. It can be observed from Table ~\ref{table_quant_exp_ewgs} that \name~does not significantly affect the performance of the full precision models as well therefore provides a strong initial checkpoint for the model to be quantized. 
\subsubsection{Model compression and quantization}
\begin{table}[t]
\caption{Model size of compressed MobileNet-V1 (in M bytes). All: all layer quantization. W/O F\&L: Quantize the model excluding the first and last layers. 2bit accuracy: EWGS 2bit weight and activation quantization accuracies on ImageNet. The model size of FP32 MobileNet-V1 is 16.1 MB} 
\label{table_model_size}
\begin{center}
\begin{sc}
\setlength{\tabcolsep}{5.5pt}
\begin{tabular}{lccc}
\toprule
\multicolumn{2}{c}{MobileNet-V1} & All  & W/O F \& L \\
\midrule
\multirow{3}{*}{Model Size (MB) }&4bit & 4.2 & 7.1   \\
&2bit & 2.2 & 5.6   \\
&1bit & 1.2 & 4.8   \\
\hline \hline
\multicolumn{2}{l}{2bit accuracy} & 55.96\%  & 59.10\% \\
\multicolumn{2}{l}{2bit accuracy with $R^2$ } & 59.05\%  & 61.25\% \\
\bottomrule

\end{tabular}

\end{sc}
\end{center}

\vskip -0.1in

\end{table}
\name~is not a model compression nor quantization method. It penalizes outlier weights during training of the base model from scratch. To evaluate the effectiveness of \name~with model compression and quantization, we apply state-of-the-art compression/quantization techniques, DKM~\cite{cho2021dkm}, LSQ~\cite{esser2019learned}, EWGS~\cite{lee2021network}, DFQ~\cite{nagel2019data}, AdaRound~\cite{nagel2020up}, SQuant~\cite{guo2022squant}, and PD-Quant~\cite{liu2023pd} to the pre-trained model with and without \name. Except EWGS~\footnote{\url{https://github.com/cvlab-yonsei/EWGS}}, SQuant~\footnote{\url{https://github.com/clevercool/SQuant}} and PD-Quant~\footnote{\url{https://github.com/hustvl/PD-Quant}}, since other works do not provide official implementation, we implement those techniques ourselves and for DFQ, and AdaRound, we used AIMET~\footnote{\url{https://quic.github.io/aimet-pages/}}.

We follow the same hyper-parameters used in the works, but \textbf{we apply compression and quantization for all layers including the first and last layers.} It is important to compress/quantize all layers including first and last layers considering computation burden at the first layer with a large convolutional filter size such as 7x7 convolutions in the first layer of ResNet and the large number of weights in the last linear layer, e.g., 1.2M of weights in the last layer of MobileNet-V1 which has 4.2M of weights in total. We have demonstrated this burden in Table \ref{table_model_size} for more clarity. Due to the outliers and wide weight ranges in the first and last layers, quantizing all layers have less accuracy than quantizing a model excluding the first and layer layers as shown in ~\cref{table_model_size}. We represent $L_\infty$ \name~as $R\_Linf$, Margin \name~as $R\_M$ and Soft-min-max \name~as $R\_Smm$ in the results.

\subsection{Model quantization}

\begin{table*}[h]
\caption{Top-1 accuracies (\%) of MobileNet-V1 and V2 on ImageNet-1K using PTQ methods with 4bit weight and 8bit activation quantization. Heavy L2: applied 10x heavy L2 regularization than baseline. KURE~\cite{kure}. Na\"ive: quantizing without any advanced PTQ techniques. DFQ~\cite{nagel2019data}, AR~\cite{nagel2020up}, SQ~\cite{guo2022squant}, PD-Q~\cite{liu2023pd}}
\vspace{-0.4cm}
\begin{center}
\begin{sc}
\setlength{\tabcolsep}{5.2pt} 
\begin{tabular}{l|c|ccccc|c|ccccc}
\hline
\multirow{2}{*}{Method} & \multicolumn{6}{c|}{MobileNet-V1} & \multicolumn{6}{c}{MobileNet-V2}\\
 & FP32 & Na\"ive & DFQ & AR & SQ & PD-Q & FP32 & Na\"ive & DFQ & AR & SQ & PD-Q\\
\midrule
Baseline &  74.12  & 2.67 &  54.06  & 70.42 & 63.85 & 71.87& 73.08 & 2.57 & 56.56 & 71.29 & 59.30 & 71.54\\
\midrule
Heavy L2 &  72.67  & 13.41 &  57.68  & 69.23 & 66.51 & 69.86& 71.00 & 4.17 & 0.09 & 68.06 & 57.30& 68.92\\
KURE &   72.50   & 53.69 &  59.21  & 60.51 & 68.84& 71.39& 72.49 & 39.21 & 62.49 & 71.68 & 66.09 & 71.88\\
\midrule
R\_Linf &  73.65  & \textbf{61.73}  & 53.00   & \textbf{72.28} & \textbf{69.69}& \textbf{72.76}& 72.64 & 59.95 & 62.39 & 71.76 & 64.02 & \textbf{72.01}\\
R\_M &  73.54  & \textbf{61.66}  &  \textbf{65.06}  & \textbf{72.29} & 68.96& \textbf{72.77}& 72.73 & \textbf{60.03} & \textbf{66.04} & \textbf{71.89} & \textbf{67.17} & 71.92\\
R\_SMM & 73.95 & 44.24 & 59.21 & 71.35 & 67.86& 72.39 & 72.81 & 36.69 & 51.46 & 71.77 & \textbf{67.16}& 71.98\\
\bottomrule

\end{tabular}
\end{sc}
\end{center}
\label{table_PTQ}
\vskip -0.3in
\end{table*}

\begin{table}[t]
\caption{Top-1 accuracies (\%) of 2bit weight and 8bit activation PTQ using PD-Quant~\cite{liu2023pd}. MNV1: MobileNet-V1, MNV2: MobileNet-V2, RN50: ResNet-50, RN101: ResNet-101. FP32 accuracy is in  \cref{table_PTQ} and \cref{table_ptq_resnet}.}
\label{table_PTQ-2bit}
\begin{center}
\begin{sc}
\begin{tabular}{l|c|c|c|c}
\toprule
Method & MNV1 & MNV2 & RN50 & RN101 \\
\midrule
Baseline    & 47.62 & 50.66 & 62.92 & 66.50\\
\midrule
R\_Linf    & \textbf{56.79} & \textbf{59.49} & 71.34 & \textbf{73.89}\\
R\_M    & 56.27 & 58.70 & \textbf{71.65} & \textbf{73.87}\\
R\_SMM    & 54.20 & 57.53 & 69.31 & 71.24\\

\bottomrule
\end{tabular}
\end{sc}
\end{center}
\vskip -0.3in
\end{table}



\subsubsection{Post-Training Quantization, PTQ with $R^2$}
We compare models trained using \name~and other weight regularization, L2, heavy L2, and KURE ~\cite{kure} and quantized using PTQ methods such as DFQ ~\cite{nagel2019data}, AdaRound ~\cite{nagel2020up}, SQaunt~\cite{guo2022squant}, and PD-Quant~\cite{liu2023pd}. There are two major techniques in DFQ, bias correction compensating bias in activation and cross-layer equalization applying scale factor per channel to make all channels in a layer have similar weight range. AdaRound adaptively rounds weights to quantization bins instead of naive nearest rounding. SQuant decomposes a layer by the Hessian-based optimization objective into three diagonal sub-items, element-
wise, kernel-wise, and output channel-wise, and then it compose the sub-items in a quantized domain with respect to
constrained absolute some of error. PD-Quant quantizes weights by comparing model prediction result before and after quantization of each layer. Unlike other PTQ, SQuant dose not require calibration dataset, so its accuracy could be less than others.

As shown in ~\cref{table_PTQ}, models trained with \name~are more quantization friendly than other regularization. As KURE makes the weight distribution uniform, it can reduce outliers as a side-effect while keeping a wide weight range. Therefore, KURE is more effective than L2 norm, but \name~shows the best accuracy as it reduces outliers as well as weight range. On the other hand, heavy L2 regularization (10x L2) makes weight ranges smaller, but it does not remove outliers, therefore prove to be ineffective here.

In comparison between FP32 accuracy of baseline models and models trained with \name, we can see there are slight accuracy regression in full-precision inference as we expected in \cref{fig_weight_dist} (right). However, after quantizing, the models trained with \name~shows better accuracies for all PTQ methods that we used in ~\cref{table_PTQ}. KURE regulates entire weights to make them uniform distribution, so it is a harsh regularization. This is the reason why FP32 accuracy with KURE is inherently less than other cases including \name~which only affects to outlier (and near-outlier) weights. Also, even considering FP32 accuracy difference, still \name~achieves better performance in terms of accuracy regression from FP32 model to quantized model, e.g., KURE regresses by 11.99\% and 1.11\% by using AdaRound and PD-Quant for MobileNet-V1, respectively (72.50\% $\rightarrow$ 60.51\% and 71.39\%), while Margin \name~regresses by 1.25\% and 0.77\% (73.54\% $\rightarrow$ 72.29\% and 72.77\%). 

Even without advanced PTQ approaches such as DFQ, AdaRound, SQuant, and PD-Quant, models trained with \name~can be reasonably quantized without any further fine-tuning. In ~\cref{table_PTQ}, Na\"ive with \name~shows significantly higher accuracy than other regularization. The models with \name~have good weight distribution already from pre-training so that they can be quantized with fairly high quantization accuracies.

We conduct further studies, lower bit PTQ (2bit weight and 8bit activation) and PTQ for larger models (ResNet50 and ResNet101) as shown in \cref{table_PTQ-2bit} and \cref{table_ptq_resnet}. From the lower bit PTQ result, we can clearly see the benefit of \name. For example, accuracy of ResNet101 trained with $L_\infty$ \name~only regressed by 5.41\% (79.30\% $\rightarrow$ 73.89\%), while the baseline model trained without \name~shows huge regression, 12.95\% (79.45\% $\rightarrow$ 66.50\%).

\begin{table}[t]
\caption{Top-1 accuracies (\%) of ResNet50 and ResNet101 on ImageNet-1K with 4bit weight and 8bit activation PTQ.}
\label{table_ptq_resnet}
\begin{center}
\begin{sc}
\begin{tabular}{l|c|c|c|c}
\toprule
 \multirow{3}{*}{Method} &\multicolumn{4}{c}{ResNet50}\\
 & FP32 & AR & SQ & PD-Q  \\
\midrule
Baseline    &  78.04 & 74.73  & 74.68 & 76.60  \\
\midrule
R\_Linf    & 77.97& 76.34  & \textbf{76.28}  & 77.50 \\
R\_M    &  78.11& \textbf{77.31} &  75.32 & \textbf{77.78} \\
R\_SMM    &  78.22& 75.61  & 75.75  & 77.21 \\
\midrule
\midrule
 \multirow{3}{*}{Method} &\multicolumn{4}{c}{ResNet101}\\
 & FP32 & AR & SQ & PD-Q  \\
\midrule
Baseline    & 79.45 & 75.18  & 75.23 &  78.16 \\
\midrule
R\_Linf    & 79.30 & \textbf{78.22}  &  76.87 & \textbf{78.61}\\
R\_M    & 79.27 & \textbf{78.19} &  \textbf{77.95} &  \textbf{78.69}\\
R\_SMM    & 79.63  & 76.71  & 77.20  &  78.49\\
\bottomrule
\end{tabular}
\end{sc}
\end{center}
\vskip -0.1in
\end{table}


\subsubsection{Quantization-Aware Training, QAT with \name}
\label{sec:qat_result}
We apply state-of-the-art quantization techniques like PACT ~\cite{choi2018pact} while training the models from scratch using \name. For other quantization aware training methods like, EWGS ~\cite{lee2021network} and LSQ ~\cite{esser2019learned} we initialize the model to pre-trained ResNet-18 (RN) ~\cite{he2016deep}, MobileNet-V1 (MN1) ~\cite{howard2017mobilenets} and MobileNet-V2 (MN2) ~\cite{sandler2018mobilenetv2} with \name.

\begin{table*}[t]
\caption{Top-1 accuracies (\%) of MobileNet-V1 and V2 on ImageNet-1K with EWGS varying model size. Weights and activations are quantized with the same 2bit. Model sizes are adjusted by varying width factor ($\alpha$ in \cite{howard2017mobilenets} and width multiplier in \cite{sandler2018mobilenetv2}). Width factor for large: 1.0, medium: 0.5, and small: 0.25.}
\label{table_quant_exp_ewgs}

\begin{center}
\begin{sc}
\setlength{\tabcolsep}{5.5pt}
\begin{tabular}{l|cc|cc|cc|cc|cc|cc}
\toprule
\multirow{3}{*}{Method}  &  \multicolumn{6}{c|}{MobileNet-V1} &  \multicolumn{6}{c}{MobileNet-V2} \\
 & \multicolumn{2}{c|}{large} & \multicolumn{2}{c|}{medium} & \multicolumn{2}{c|}{small}  & \multicolumn{2}{c|}{large} & \multicolumn{2}{c|}{medium} & \multicolumn{2}{c}{small} \\
 & FP32 & 2bit & FP32 & 2bit & FP32 & 2bit & FP32 & 2bit & FP32 & 2bit & FP32 & 2bit\\
\midrule
Baseline    & 74.12 & 55.96 &66.51 & 38.77 & 55.43 & 20.85 & 73.08 & 53.93  & 65.55 & 39.25 &  53.90 & 27.82\\
KURE    &  72.50 & 57.80& 63.99 & 39.37 & 52.83 & 22.16 & 72.49 & 53.97 & 64.64 & 38.51 &  52.80 & 26.86\\
\midrule
R\_Linf    &  73.65 & \textbf{59.05} & 65.68 & 41.22 &  53.48 & \textbf{26.04}& 72.64 & 56.35 & 65.09 & 42.55 &  53.40  & \textbf{30.56}\\
R\_M    & 73.54 & 58.41 & 65.83 & \textbf{42.61} &  53.30 & 24.35& 72.73 & \textbf{57.36} & 64.86 & \textbf{43.78} &  52.82 & 29.61\\


\bottomrule
\end{tabular}
\end{sc}
\end{center}
\vskip -0.2in
\end{table*}
\cref{table_quant_exp_ewgs} shows 2 bit weight and activation quantization result of MobileNet-V1 and V2 using EWGS ~\cite{lee2021network} with various regularization such as without \name~(only with L2 norm), KURE ~\cite{kure}, and our \name. For both the models with varying model sizes, \name~outperforms the models trained with only L2 norm or KURE. Without \name, accuracy of MobileNet-V1 2bit quantization using EWGS declines from 59.10\% to 55.96\% when we quantize all layers including the first and last layers as shown previously in \cref{table_model_size}. This is because the first and last layers have wide weight ranges and many outliers as shown in \cref{fig:weight_dist_mnv2}. Our approach effectively reduces the outliers in the first and last layer which enables the 2bit quantized model to achieve similar accuracy to the case with original EWGS results where the first and last layers of the model remain in FP32.

\begin{table}[t]
\caption{Top-1 accuracies (\%) of ResNet18 with various QAT methods. Weights and activations are quantized with the same bit (2W2A: 2bit, 4W4A: 4bit).}
\label{table_quant_exp}
\begin{center}
\begin{sc}
\begin{tabular}{l|c|c|c|c}
\toprule
\multirow{2}{*}{Method} &\multicolumn{4}{c}{2W2A EWGS}\\
 & FP32 & PACT & LSQ & EWGS \\
\midrule
Baseline    & 69.76 & 51.97  & 58.33 &  65.42 \\
\midrule
R\_Linf    & 70.15 & 55.26  & 62.23  & \textbf{65.72}\\
R\_M    & 70.08 & \textbf{56.24} & 62.25  & 64.27 \\
R\_SMM    & 69.84 & 55.64  & \textbf{62.47}  & 64.94 \\
\midrule
\midrule
\multirow{2}{*}{Method} &\multicolumn{4}{c}{4W4A EWGS}\\
 & FP32 & PACT & LSQ & EWGS \\
\midrule
Baseline    & 69.76 & 66.90  & 69.90 &  70.19 \\
\midrule
R\_Linf    & 70.15 & 68.45  & 69.55  & 70.17\\
R\_M    & 70.08 & 68.30 & 69.56  & 69.77 \\
R\_SMM    & 69.84 & 68.36  & 69.45  & 69.80 \\
\bottomrule
\end{tabular}
\end{sc}
\end{center}
\vskip -0.1in
\end{table}

As shown in \cref{table_quant_exp}, \name~helps the quantization techniques in improving their accuracy, especially for extremely low bit quantization such as at 2 bit while it shows similar accuracies with 4 bit. For example, all \name es improve 2 bit quantization accuracy with LSQ to over than 62\% from 58\%, but there is no noticeable difference in 4 bit LSQ accuracies with and without $R^2$. The reason why \name~would not help much for higher bit like 4 bit quantization is that QAT can effectively represent outliers using many bits as we expected in ~\cref{fig_weight_dist} (right).


Interestingly, soft-min-max \name~does not seem to be as good as $L_\infty$ or margin \name~for quantization. As discussed in \cref{sec:smm}, soft-min-max \name~allows us to have an asymmetric weight distribution so that it would be more effective for model compression instead of symmetric model quantization. 

\subsection{Model compression}
\label{sec:model_compression}

\begin{table}[t]
\caption{Top-1 accuracies(\%) of compression using DKM with ResNet18 (RN), MobileNet-V1 (MN1) on ImageNet varying compression bit. FP32 accuracy is in \cref{table_quant_exp_ewgs} and \cref{table_quant_exp}.} 
\label{table_comp_exp}
\begin{center}
\begin{sc}
\begin{tabular}{llcccc}
\toprule
Method & $R^{2}$ & RN & MN1\\

\midrule
\multirow{4}{*}{DKM 1-bit, 1-dim}    & Baseline        & 58.97 & 45.54\\\cmidrule{2-4}
                        & R\_Linf    & 59.52 & 49.74 \\
                        & R\_M    &  \textbf{59.70} & 47.21\\
                        & R\_SMM    & 59.27 & \textbf{52.58} \\
                        \midrule
\multirow{4}{*}{DKM 2-bit, 1-dim}    & Baseline     & 67.64 & 65.95 \\\cmidrule{2-4}
                                     & R\_Linf  & 68.53 & 67.06 \\
                                     & R\_M     & 68.33 & 67.50 \\
                                     & R\_SMM   & \textbf{68.63} & \textbf{67.62} \\
                        \midrule
\multirow{4}{*}{DKM 4-bit, 1-dim}    
                        & Baseline     & 70.22 & 69.29 \\\cmidrule{2-4}
                        & R\_Linf    & 70.34 & 69.43 \\
                        & R\_M    & 70.33 & 68.52 \\
                        & R\_SMM    & \textbf{70.52} & \textbf{69.63} \\
\bottomrule
\end{tabular}
\end{sc}
\end{center}
\vskip -0.1in
\end{table}

\begin{table}[t]
\caption{Top-1 accuracies(\%) of compression using multi-dimensional DKM and \name~with ResNet18 (RN), MobileNet\_V1 (MN1) on ImageNet.} 
\label{table_comp_exp_multi}
\vskip -0.15in
\begin{center}
\begin{sc}
\begin{tabular}{llcccc}
\toprule
Method & $R^{2}$ & RN & MN1 \\
\midrule
\multirow{2}{*}{DKM 2-bit, 2-dim}    & Baseline        & 63.52 & 48.16 \\\cmidrule{2-4}
                        & R\_SMM    & \textbf{64.64} & \textbf{53.99}\\
                        \midrule
\multirow{2}{*}{DKM 4-bit, 4-dim}    & Baseline        & 64.89 & 58.55\\\cmidrule{2-4}
                        & R\_SMM    & \textbf{66.10} & \textbf{60.05} \\
\bottomrule
\end{tabular}
\end{sc}
\end{center}
\vskip -0.1in
\end{table}

We evaluate the effectiveness of \name~for compression with the state-of-the-art compression technique, DKM \cite{cho2021dkm}, for ResNet-18 and MobileNet-V1. The bit-dim ratio, $\frac{b}{d}$ is an important factor in the DKM algorithm which effectively defines the kind of compression a DKM palettized model would see. We ran these experiments for both scalar and vector palettization. For scalar palettization($dim=1$) we ran 1 bit, 2 bit and 4 bit compression. These experiments would yield roughly 32x, 16x and 8x compressed models respectively. \cref{table_comp_exp} shows that \name~significantly improves accuracy from original scalar palettized DKM 1 bit, 2 bit and 4 bit models. As we discussed, there is no significant difference for higher bit compression like 4 bit because many bit compression can also cover outliers even without \name.

We also expand the application of \name~to vector palettization($dim>1$) DKM \cite{cho2021dkm} as demonstrated in \cref{table_comp_exp_multi}. For these experiments, we kept the effective bit-dim ratio, $\frac{b}{d}$ equivalent to 1 so as to see variation across the most heavily compressed model which would be close to 32x compressed. Since a vector palettized model will require range constraining for all dimensions, we applied multi-dimensional \name~for all layers that would be palettized during compression. For vector palettized ResNet-18 there is an average absolute improvement of $>1\%$ using models trained with \name, and for vector palettized MobileNet-V1, the gain ranges from 5\% to 3\%. 

Finally we also validated that \name~scales to other domains as well by applying it in compressing MobileBERT \cite{sun2020mobilebert}. For Question Answering (QNLI) \cite{rajpurkar2016squad} using MobileBERT, \name~improved the performance of the model by 2\% absolute as demonstrated in Table \ref{exp_mobilebert}. Note that we applied \name~to a QNLI fine-tuning task based on a pre-trained MobileBERT \cite{wolf2020transformers}. It might be necessary to apply $R^2$ to the entire training task of MobileBERT from scratch so that \name~would have more chances to get effective weight distribution for model compression. Through these experiments across a variety of tasks(Image Classification, Question Answering etc.) we can also see that the application of the proposed \name~is task-invariant and yields solid results across domains.

\begin{table}[h]
\caption{Question-answering NLI (QNLI) accuracies of MobileBERT using single dimension DKM}
\label{exp_mobilebert}
\vskip -0.15in
\begin{center}
\begin{sc}
\begin{tabular}{lccc}
\toprule
Method &Pre-train & 1-bit & 2-bit \\
\midrule
DKM baseline    & 90.41 & 61.34 & 80.12 \\
\midrule
DKM + R\_Linf  & 90.66 & \textbf{63.17} & \textbf{82.13}        \\
DKM + R\_M  & 89.09 & 61.80 & 80.98        \\
DKM + R\_SMM & 90.83 & 61.49 & 80.87        \\
\bottomrule
\end{tabular}
\end{sc}
\end{center}
\vskip -0.1in
\end{table}

\subsection{Strength of \name~\& comparison between \name}
\label{sec:model_compression}


\begin{figure}[t!]
\vskip 0.0in
\begin{center}
\includegraphics[trim={9 0 12 0},clip, width=1\linewidth]{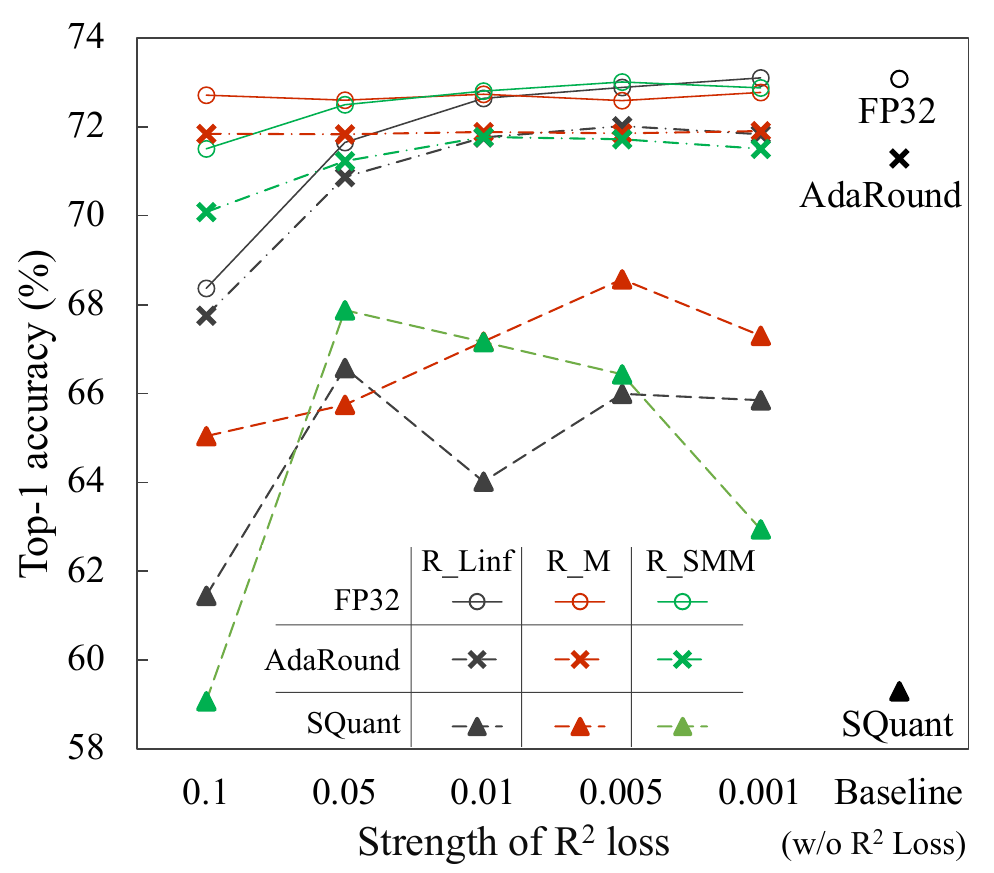}
\vskip -0.05in
\caption{MobileNet-V2 result varying strength of \name. 4bit weight and 8bit activation PTQ with AdaRound and SQuant.}
\label{fig:r2_strength}
\end{center}
\vskip -0.1in
\end{figure}

In this paper, we propose three different \name, $L_\infty$, Margin, and Soft-Min-Max. As we discussed in \cref{sec:smm} and \cref{sec:qat_result}, Soft-Min-Max \name~would be more effective for model compression as it allows to have asymmetric weight distribution while other \name~look better for symmetric quantization. To compare $L_\infty$ and Margin \name~and see impact of strength of \name, we conduct a further ablation study varying strength of \name as shown in \cref{fig:r2_strength}. While $L_\infty$ \name~shows somewhat fluctuated quantization performance with respect to its strength, quantization accuracy from a model trained with Margin \name~is more consistent and better than $L_\infty$. $L_\infty$ \name~is naive and simple to limit the weight range. Margin \name~is advanced loss from $L_\infty$ as it applies learnable margin threshold. We think that the learnable margin parameter makes more stable full-precision pre-training while $L_\infty$ always penalizes only one outlier per iteration. Therefore, Margin (as well as Soft-min-max) \name~shows more stable full-precision accuracy before quantization. We would recommend to use Margin \name~for symmetric quantization if it is hard to find proper strength for $L_\infty$ \name.

\section{Conclusion}
In this paper, we introduced \fname~as an effective technique to get rid of outliers in the weight distribution during training. This serves as a good initialization for state of the art post training quantization, quantization aware training and compression strategies, therefore can be coupled with any of them. This helps to augment the accuracy gained from such techniques and is invariant to the quantization or compression algorithm. With the three proposed formulations of \name~we wanted to demonstrate that there can multiple approaches of defining \name~and the final metric depends on these formulations. We also demonstrated how the proposed method converts a wide weight range distribution to a more densely-packed distribution for model quantization. While full-precision accuracy with \name~can be slightly regressed as it penalizes outlier weights, \name~significantly improves quantization and compression accuracy, especially for lower bit.
\section{Impact Statement}
Deep learning models play a pivotal role in multiple applications today. With time these models have become larger and larger in search of higher accuracy. However, deploying such models on edge devices is challenging because of their size. We introduce a simple yet effective algorithm to train models suitable for ultra low bit quantization and compression (e.g., 1 and 2 bit). We hope that the community will adopt this method to create compute-efficient models, resulting in reduced deployment costs. There are many potential societal consequences of our work, none which we feel must be specifically highlighted here.

\nocite{langley00}

\bibliography{icml2024}

\begin{thebibliography}{24}
\providecommand{\natexlab}[1]{#1}
\providecommand{\url}[1]{\texttt{#1}}
\expandafter\ifx\csname urlstyle\endcsname\relax
  \providecommand{\doi}[1]{doi: #1}\else
  \providecommand{\doi}{doi: \begingroup \urlstyle{rm}\Url}\fi

\bibitem[Bengio et~al.(2013)Bengio, L{\'e}onard, and Courville]{bengio2013estimating}
Bengio, Y., L{\'e}onard, N., and Courville, A.
\newblock Estimating or propagating gradients through stochastic neurons for conditional computation.
\newblock \emph{arXiv preprint arXiv:1308.3432}, 2013.

\bibitem[Cho et~al.(2021)Cho, Alizadeh-Vahid, Adya, and Rastegari]{cho2021dkm}
Cho, M., Alizadeh-Vahid, K., Adya, S., and Rastegari, M.
\newblock Dkm: Differentiable k-means clustering layer for neural network compression.
\newblock In \emph{International Conference on Learning Representations}, 2021.

\bibitem[Choi et~al.(2018)Choi, Wang, Venkataramani, Chuang, Srinivasan, and Gopalakrishnan]{choi2018pact}
Choi, J., Wang, Z., Venkataramani, S., Chuang, P. I.-J., Srinivasan, V., and Gopalakrishnan, K.
\newblock Pact: Parameterized clipping activation for quantized neural networks.
\newblock \emph{arXiv preprint arXiv:1805.06085}, 2018.

\bibitem[Deng et~al.(2009)Deng, Dong, Socher, Li, Li, and Fei-Fei]{deng2009imagenet}
Deng, J., Dong, W., Socher, R., Li, L.-J., Li, K., and Fei-Fei, L.
\newblock Imagenet: A large-scale hierarchical image database.
\newblock In \emph{2009 IEEE conference on computer vision and pattern recognition}, pp.\  248--255. Ieee, 2009.

\bibitem[Dong et~al.()Dong, Yao, Arfeen, Gholami, Mahoney, and Keutzer]{hawq}
Dong, Z., Yao, Z., Arfeen, D., Gholami, A., Mahoney, M.~W., and Keutzer, K.
\newblock Hawq-v2: Hessian aware trace-weighted quantization of neural networks.
\newblock In \emph{Proceedings of the 34th International Conference on Neural Information Processing Systems}, NIPS'20.

\bibitem[Esser et~al.(2019)Esser, McKinstry, Bablani, Appuswamy, and Modha]{esser2019learned}
Esser, S.~K., McKinstry, J.~L., Bablani, D., Appuswamy, R., and Modha, D.~S.
\newblock Learned step size quantization.
\newblock In \emph{International Conference on Learning Representations}, 2019.

\bibitem[Guo et~al.(2022)Guo, Qiu, Leng, Gao, Zhang, Liu, Yang, Zhu, and Guo]{guo2022squant}
Guo, C., Qiu, Y., Leng, J., Gao, X., Zhang, C., Liu, Y., Yang, F., Zhu, Y., and Guo, M.
\newblock {SQ}uant: On-the-fly data-free quantization via diagonal hessian approximation.
\newblock In \emph{International Conference on Learning Representations}, 2022.
\newblock URL \url{https://openreview.net/forum?id=JXhROKNZzOc}.

\bibitem[Han et~al.(2016)Han, Mao, and Dally]{HanMD15}
Han, S., Mao, H., and Dally, W.~J.
\newblock Deep compression: Compressing deep neural network with pruning, trained quantization and huffman coding.
\newblock In Bengio, Y. and LeCun, Y. (eds.), \emph{4th International Conference on Learning Representations, {ICLR} 2016, San Juan, Puerto Rico, May 2-4, 2016, Conference Track Proceedings}, 2016.

\bibitem[Han et~al.(2021)Han, Li, Liu, Tian, and Shan]{binregularization}
Han, T., Li, D., Liu, J., Tian, L., and Shan, Y.
\newblock Improving low-precision network quantization via bin regularization.
\newblock In \emph{2021 IEEE/CVF International Conference on Computer Vision (ICCV)}, pp.\  5241--5250, Los Alamitos, CA, USA, oct 2021. IEEE Computer Society.

\bibitem[He et~al.(2016)He, Zhang, Ren, and Sun]{he2016deep}
He, K., Zhang, X., Ren, S., and Sun, J.
\newblock Deep residual learning for image recognition.
\newblock In \emph{Proceedings of the IEEE conference on computer vision and pattern recognition}, pp.\  770--778, 2016.

\bibitem[Howard et~al.(2017)Howard, Zhu, Chen, Kalenichenko, Wang, Weyand, Andreetto, and Adam]{howard2017mobilenets}
Howard, A.~G., Zhu, M., Chen, B., Kalenichenko, D., Wang, W., Weyand, T., Andreetto, M., and Adam, H.
\newblock Mobilenets: Efficient convolutional neural networks for mobile vision applications.
\newblock \emph{arXiv preprint arXiv:1704.04861}, 2017.

\bibitem[Langley(2000)]{langley00}
Langley, P.
\newblock Crafting papers on machine learning.
\newblock In Langley, P. (ed.), \emph{Proceedings of the 17th International Conference on Machine Learning (ICML 2000)}, pp.\  1207--1216, Stanford, CA, 2000. Morgan Kaufmann.

\bibitem[Lee et~al.(2021)Lee, Kim, and Ham]{lee2021network}
Lee, J., Kim, D., and Ham, B.
\newblock Network quantization with element-wise gradient scaling.
\newblock In \emph{Proceedings of the IEEE/CVF Conference on Computer Vision and Pattern Recognition}, pp.\  6448--6457, 2021.

\bibitem[Liu et~al.(2023)Liu, Niu, Yuan, Yang, Wang, and Liu]{liu2023pd}
Liu, J., Niu, L., Yuan, Z., Yang, D., Wang, X., and Liu, W.
\newblock Pd-quant: Post-training quantization based on prediction difference metric.
\newblock In \emph{Proceedings of the IEEE/CVF Conference on Computer Vision and Pattern Recognition}, pp.\  24427--24437, 2023.

\bibitem[Nagel et~al.(2019)Nagel, Baalen, Blankevoort, and Welling]{nagel2019data}
Nagel, M., Baalen, M.~v., Blankevoort, T., and Welling, M.
\newblock Data-free quantization through weight equalization and bias correction.
\newblock In \emph{Proceedings of the IEEE/CVF International Conference on Computer Vision}, pp.\  1325--1334, 2019.

\bibitem[Nagel et~al.(2020)Nagel, Amjad, Van~Baalen, Louizos, and Blankevoort]{nagel2020up}
Nagel, M., Amjad, R.~A., Van~Baalen, M., Louizos, C., and Blankevoort, T.
\newblock Up or down? adaptive rounding for post-training quantization.
\newblock In \emph{International Conference on Machine Learning}, pp.\  7197--7206. PMLR, 2020.

\bibitem[Rajpurkar et~al.(2016)Rajpurkar, Zhang, Lopyrev, and Liang]{rajpurkar2016squad}
Rajpurkar, P., Zhang, J., Lopyrev, K., and Liang, P.
\newblock Squad: 100,000+ questions for machine comprehension of text.
\newblock \emph{arXiv preprint arXiv:1606.05250}, 2016.

\bibitem[Sandler et~al.(2018)Sandler, Howard, Zhu, Zhmoginov, and Chen]{sandler2018mobilenetv2}
Sandler, M., Howard, A., Zhu, M., Zhmoginov, A., and Chen, L.-C.
\newblock Mobilenetv2: Inverted residuals and linear bottlenecks.
\newblock In \emph{Proceedings of the IEEE conference on computer vision and pattern recognition}, pp.\  4510--4520, 2018.

\bibitem[Shkolnik et~al.(2020)Shkolnik, Chmiel, Banner, Shomron, Nahshan, Bronstein, and Weiser]{kure}
Shkolnik, M., Chmiel, B., Banner, R., Shomron, G., Nahshan, Y., Bronstein, A., and Weiser, U.
\newblock Robust quantization: One model to rule them all.
\newblock In \emph{Proceedings of the 34th International Conference on Neural Information Processing Systems}, NIPS'20, Red Hook, NY, USA, 2020. Curran Associates Inc.
\newblock ISBN 9781713829546.

\bibitem[Sun et~al.(2020)Sun, Yu, Song, Liu, Yang, and Zhou]{sun2020mobilebert}
Sun, Z., Yu, H., Song, X., Liu, R., Yang, Y., and Zhou, D.
\newblock Mobilebert: a compact task-agnostic bert for resource-limited devices.
\newblock \emph{arXiv preprint arXiv:2004.02984}, 2020.

\bibitem[Wang et~al.(2019)Wang, Liu, Lin, Lin, and Han]{haq}
Wang, K., Liu, Z., Lin, Y., Lin, J., and Han, S.
\newblock Haq: Hardware-aware automated quantization with mixed precision.
\newblock pp.\  8604--8612, 06 2019.
\newblock \doi{10.1109/CVPR.2019.00881}.

\bibitem[Wolf et~al.(2020)Wolf, Debut, Sanh, Chaumond, Delangue, Moi, Cistac, Rault, Louf, Funtowicz, et~al.]{wolf2020transformers}
Wolf, T., Debut, L., Sanh, V., Chaumond, J., Delangue, C., Moi, A., Cistac, P., Rault, T., Louf, R., Funtowicz, M., et~al.
\newblock Transformers: State-of-the-art natural language processing.
\newblock In \emph{Proceedings of the 2020 conference on empirical methods in natural language processing: system demonstrations}, pp.\  38--45, 2020.

\bibitem[Yin et~al.(2019)Yin, Lyu, Zhang, Osher, Qi, and Xin]{yinl2019}
Yin, P., Lyu, J., Zhang, S., Osher, S.~J., Qi, Y., and Xin, J.
\newblock Understanding straight-through estimator in training activation quantized neural nets.
\newblock In \emph{7th International Conference on Learning Representations, {ICLR} 2019, New Orleans, LA, USA, May 6-9, 2019}. OpenReview.net, 2019.

\bibitem[Zhou et~al.(2016)Zhou, Wu, Ni, Zhou, Wen, and Zou]{zhou2016dorefa}
Zhou, S., Wu, Y., Ni, Z., Zhou, X., Wen, H., and Zou, Y.
\newblock Dorefa-net: Training low bitwidth convolutional neural networks with low bitwidth gradients.
\newblock \emph{arXiv preprint arXiv:1606.06160}, 2016.

\end{thebibliography}
\bibliographystyle{icml2024}

\end{document}